\documentclass{article} \pdfpagewidth=8.5in
\usepackage{arxiv}

\usepackage{times}
\usepackage{soul}
\usepackage{url}
\usepackage[hidelinks]{hyperref}
\usepackage[utf8]{inputenc}
\usepackage[small]{caption}
\usepackage{graphicx}
\usepackage{amsmath}
\usepackage{amsthm}
\usepackage{booktabs}
\usepackage{algorithm}
\usepackage{algcompatible}
\usepackage[noend]{algpseudocode}
\usepackage[switch]{lineno}
\usepackage{varwidth}
\usepackage[symbol]{footmisc}
\usepackage{color}
\usepackage{xcolor}
\usepackage{xspace}

\newcommand{\pw}{PushWorld\xspace}

\newcommand{\pwplayurl}{\url{https://deepmind-pushworld.github.io/play/}} 
\newcommand{\pwcodeurl}{\url{https://github.com/deepmind/pushworld/}} 

\algnewcommand{\LineComment}[1]{\State \(\triangleright\) \text{#1}}

\title{\pw: A benchmark for manipulation planning with tools and movable obstacles}

\author{
Ken Kansky$^1$\and
Skanda Vaidyanath$^{2*}$\and
Scott Swingle$^3$\and
Xinghua Lou$^3$\and
Miguel L\'azaro-Gredilla$^3$\and
Dileep George$^3$\\
\affiliations
$^1$Intrinsic Innovation\\
$^2$Stanford University\\
$^3$DeepMind\\
\emails
\pwplayurl
}

\begin{document}

\maketitle

\def\thefootnote{*}\footnotetext{During author's internship at DeepMind.}

\renewcommand*{\thefootnote}{\arabic{footnote}}
\setcounter{footnote}{0}

\begin{abstract}

While recent advances in artificial intelligence have achieved human-level performance in environments like Starcraft and Go, many physical reasoning tasks remain challenging for modern algorithms. To date, few algorithms have been evaluated on physical tasks that involve manipulating objects when movable obstacles are present and when tools must be used to perform the manipulation. To promote research on such tasks, we introduce \pw, an environment with simplistic physics that requires manipulation planning with both movable obstacles and tools. We provide a benchmark of more than 200 \pw puzzles in PDDL and in an OpenAI Gym environment. We evaluate state-of-the-art classical planning and reinforcement learning algorithms on this benchmark, and we find that these baseline results are below human-level performance. We then provide a new classical planning heuristic that solves the most puzzles among the baselines, and although it is 40 times faster than the best baseline planner, it remains below human-level performance.

\end{abstract}

\section{Introduction}

\begin{figure*}[htbp!]
\centering
\includegraphics[width=\textwidth]{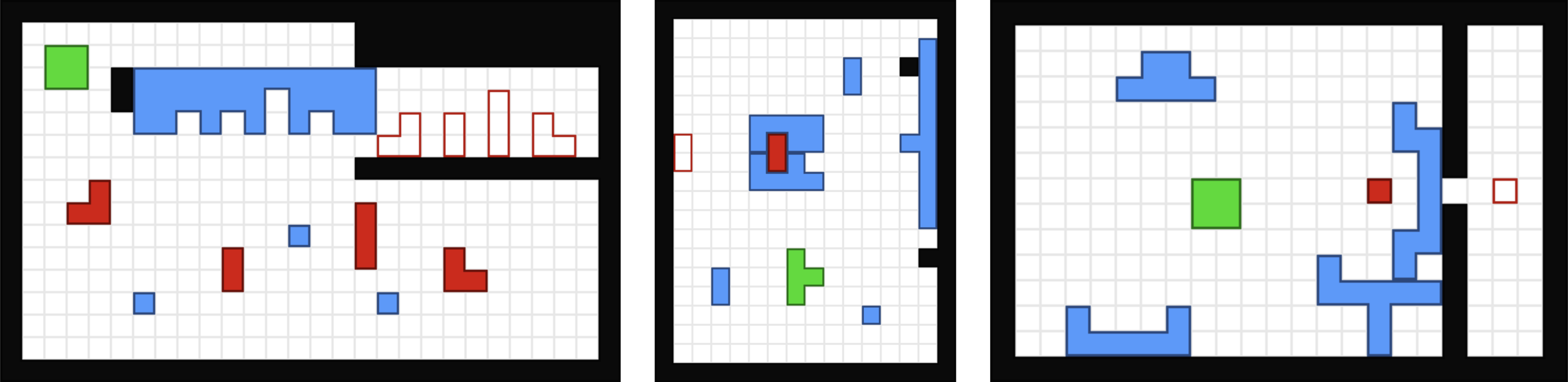}
\caption{Example \pw puzzles. Black regions are immovable walls, and the green object is the agent. The agent can push blue and red objects, which can transitively push other blue and red objects. To solve a puzzle, all red objects must move to their outlined goal positions. \textbf{(left)} An obstacle prevents the agent from directly pushing the objects into their goal positions. Instead, the agent must arrange all red objects in their corresponding slots in the obstacle and then use the obstacle as a tool to simultaneously push them into place. There are three distracting objects that do not help achieve the goals. \textbf{(center)} The goal object is enclosed by two obstacles that prevent pushing the object into its outlined position. The agent's shape is unable to push the obstacles off of the goal object, so the agent must push the enclosure onto the peg of the object on the right wall and move the peg upward to pull apart the enclosure. All other objects are useless distractors. \textbf{(right)} A challenging puzzle. The agent must remove the obstacle that blocks the gap in the wall. The agent is too large to pull the obstacle away from the wall, so it must use the lower left U-shaped object as a tool that hooks behind the obstacle to help pull it away. However, the lower right object is an obstacle in the way of the tool, so the agent must first remove this second obstacle. Finally, the agent cannot fit through the gap in the wall, so the agent must use the lower right object as a tool to push the red object through the gap to its goal.
}
\label{fig:banner}
\end{figure*}

In everyday tasks, people manipulate objects to achieve their goals, like grasping a coffee cup from a shelf and rotating it to lift it over other cups. If there is not enough space to lift the cup over others, people will push the other cups out of the way, or if the cup is out of reach, people might use a spoon as a tool to pull the handle of the cup within reach. This task is an example of manipulation planning in the presence of tools and movable obstacles. People are adept at solving such tasks, but reaching human performance on these tasks remains an active research area in artificial intelligence and robotics \cite{garrett2020pddlstream,stilman2008planning,toussaint2018differentiable}.

One research approach is to construct simplified environments to simulate real-world tasks, then develop algorithms in these simplified environments, and then generalize these algorithms for use in the real world. For example, grid-based environments such as Sokoban have been extensively used for evaluating physical planning algorithms \cite{dor1999sokoban,thrun2002probabilistic}, and these algorithms are now applied in real-world robotic tasks \cite{garrett2020pddlstream}. Indeed, environments like Sokoban have contributed to the development of a variety of intelligent algorithms, including tree search with pattern database heuristics \cite{haslum2007domain} and using features as a high-level road map for search \cite{shoham2020fess}. The deep reinforcement learning community also continues to develop Sokoban solvers in both model-free \cite{guez2019investigation} and model-based \cite{hamrick2020role} settings.

Despite its popularity as a planning benchmark, Sokoban does not involve manipulation planning or tool use. Moreover, existing environments for robotic manipulation planning typically do not include both tools and movable obstacles \cite{dogar2011framework,simeon2004manipulation,stilman2008planning,toussaint2018differentiable}. Therefore, we introduce \pw, a simplified environment for testing manipulation planning in the presence of both tools and movable obstacles. As shown in Figure~\ref{fig:banner}, \pw is a two-dimensional physical environment with finite, discrete states. Like Sokoban, it is fully observable, deterministic, and static. However, \pw additionally allows objects to have diverse shapes and sizes, and multiple objects can be pushed simultaneously. These differences from Sokoban introduce many new challenges: a \pw puzzle may require pushing multiple obstacles out of the way simultaneously, or it may require assembling a tool to push another object. Section \ref{pw_env} elaborates on the challenges that \pw offers.

\pw is also usable as a sparse-reward and hierarchical reinforcement learning (RL) environment that is simpler than many existing environments that involve robots \cite{her} or continuous action spaces \cite{todorov2012mujoco}. Three-dimensional RL environments have been created to test tool use \cite{team2021open}, but \pw demonstrates that non-trivial tool use problems are plentiful in only two dimensions.

Our contributions are as follows: First, we introduce \pw, a simple physical environment suitable for research in classical planning, reinforcement learning, combined task and motion planning for robotics, and cognitive science. Second, we benchmark several state-of-the-art classical planning and deep reinforcement learning algorithms on \pw as baseline evaluations. Third, we provide a novel planning heuristic that outperforms all baselines. Finally, we provide an open-source package\footnote[1]{\pwcodeurl} that includes an OpenAI Gym environment, a benchmarking dataset of 223 hand-designed \pw puzzles with scripts to generate PDDL and SAS+ \cite{backstrom1995complexity}, and the novel planning heuristic.

\section{The \pw Environment}
\label{pw_env}

As shown in Figure~\ref{fig:banner}, \pw is a simplified two-dimensional physical environment in which the goal is to push one or more objects into target positions. Objects have rigid shapes that occupy sets of discrete positions, which cannot intersect with the occupied positions of other objects. Objects can translate but cannot rotate.

One object is designated as the \textit{agent}, which can perform one of four actions at each discrete time step: moving up, down, left, or right by a single discrete position. The agent can push other objects in the direction of its movement, and there is no limit on the number of objects that the agent can push simultaneously. Any object can also push any other, so the agent can use objects as tools to indirectly push other objects. Objects have no momentum, so they immediately stop moving after the agent pushes them, as if all objects rest on a high-friction surface. Immovable walls also occupy discrete positions, and actions are forbidden that would push any object into a wall's position. Figure~\ref{fig:banner} illustrates three example puzzles, and Appendix \ref{appendix.puzzle_def} provides a formal definition of a \pw puzzle.

In spite of its simplified physics, solving \pw puzzles requires diverse skills:

\begin{itemize}
    \item \textbf{Path Planning} involves finding a collision-free path to move an object from one position to another, considering walls and object shapes.
    \item \textbf{Manipulation Planning} involves exploring alternative ways to push an object along a desired path. Some puzzles require the agent to preemptively position itself so that it can move from pushing in one direction to pushing in a different direction in the future.
    \item \textbf{Movable Obstacles} require the agent to decide between finding a path around an obstacle or moving the obstacle out of the way. Some obstacles introduce choices in which freeing one path blocks another, and choices can be irreversible. Due to object shapes and the inability to pull objects, obstacles can also be ``parasitic": once pushed against an object, the agent can never separate the obstacle from the object.
    \item \textbf{Tool Use}. To move an object into a desired position, the agent may need to use one or more objects as tools to indirectly push the target object: the agent pushes a tool object, which simultaneously pushes the target or another tool. In some puzzles the agent must assemble a tool by pushing multiple objects together.
    \item \textbf{Prioritizing Multiple Goals}. Moving an object into its goal position too soon may prevent achieving another goal, and some puzzles require achieving multiple goals simultaneously by preemptively arranging objects and pushing them all at once.
\end{itemize}

Within a single puzzle, an object can change its role over time, sometimes functioning as a tool, as an obstacle, or as a manipulation target. \pw therefore requires flexibly applying the skills above to adapt to changing situations.

As a reinforcement learning environment, \pw also poses several challenges:

\begin{itemize}
    \item \textbf{Sparse rewards:} The agent only receives a reward when it pushes an object into its goal position.
    Compared to Sokoban, \pw puzzles tend to require more actions to achieve all goals, which makes learning from rewards more challenging.
    \item \textbf{Credit assignment:} Judging which actions contributed to causing an outcome can hasten RL training \cite{hca} and can improve exploration \cite{yan2022deep}.
    In some \pw puzzles it is possible to reach states from which it is no longer possible to achieve all goals, so RL agents can avoid these states by identifying their causes. A model-based approach can answer counterfactual questions that assist with credit assignment \cite{cca}.
    \item \textbf{Exploration:} Efficient exploration is one strategy to counteract sparse rewards \cite{https://doi.org/10.48550/arxiv.1606.01868,NEURIPS2018_a2802cad}.
    An intelligent exploration algorithm can identify which parts of the state space to explore and which to ignore \cite{https://doi.org/10.48550/arxiv.2008.02790}.
    The agent must also explore safely to avoid states from which goals are no longer achievable \cite{https://doi.org/10.48550/arxiv.2106.04480}.
    \item \textbf{Hierarchy:} Humans can decompose \pw puzzles into subgoals to remove obstacles, retrieve tools, navigate to desired positions, and more. This decomposition suggests that hierarchical approaches \cite{florensa2017stochastic,https://doi.org/10.48550/arxiv.2203.00054,https://doi.org/10.48550/arxiv.1906.05862,https://doi.org/10.48550/arxiv.1805.08296} may be necessary to achieve human-level performance.
\end{itemize}

\section{Related Environments}

\begin{table*}
  \centering
  \begin{tabular}{lccccc}
  {} & Sokoban & Sliding Block Puzzles & Robotic Manipulation & Grid Path Planning & \pw \\
  \hline
  Varied Object Shapes & No & Yes & Yes & Yes & Yes \\
  Path Planning & Yes & No & Yes & Yes & Yes \\
  Manipulation Planning & No & No & Yes & No & Yes \\
  Movable Obstacles & Yes & Yes & Sometimes & No & Yes \\
  Tool Use & No & No & Sometimes & No & Yes \\
  Multiple Goals & Yes & Yes & Sometimes & No & Yes \\
  Simplified Physics & Yes & Yes & No & Yes & Yes \\
  \end{tabular}
  \caption{Properties of \pw vs. related environments as a planning benchmark.}
  \label{related_envs_table}
\end{table*}

\subsection{Sokoban}

A Sokoban puzzle (Fig.~\ref{fig:related-envs-samples}A) requires an agent to push one or more boxes into goal positions. Unlike \pw, the agent can only push one box at a time, and all boxes have the same shape. By allowing objects to have different shapes and allowing the agent to push multiple objects at once, \pw adds complexity to path planning and obstacle interactions, and it introduces the use of tools. Like Sokoban, \pw puzzles include the potential for deadlocks, in which action effects are irreversible, and for conflicts, in which achieving a goal in the wrong order may prevent achieving other goals.

\subsection{Sliding Block Puzzles}

Sliding block puzzles involve sliding one or more 2D objects into goal positions, and well-known examples include Klotski (Fig.~\ref{fig:related-envs-samples}B) and the fifteen puzzle. Objects can have varying shapes, and actions can directly move all objects. Challenging sliding block puzzles typically confine all objects without much free space to move. In \pw, actions can directly move only one object, the agent, so limited free space is not necessary to construct challenging puzzles.

\subsection{Robotic Manipulation Among Movable Obstacles}

A robotic manipulation task involves moving one or more objects into goal states, typically by executing alternating phases of \textit{transit}, during which the robot moves without moving any other object, and \textit{transfer}, during which the robot controls the movement of one or more objects \cite{simeon2004manipulation}. Tasks may require moving objects out of the way to make space for moving the robot or for manipulating another object \cite{stilman2008planning,stilman2007manipulation}, and in some tasks the robot must push objects out of the way because they are too heavy to lift or too large to grasp \cite{dogar2011framework}. While \pw has a smaller state space than these robotic tasks, \pw nevertheless requires similar planning capabilities to move obstacles and to manipulate objects via pushing. Some robotic tasks involve tool use \cite{toussaint2018differentiable} (Fig.~\ref{fig:related-envs-samples}C), but robotic tasks involving both tool use and movable obstacles have received less attention. \pw therefore presents a novel integration of planning challenges.

\subsection{Grid-Based Path Planning}

\begin{figure}[htbp!]
\centering
\includegraphics[width=\linewidth]{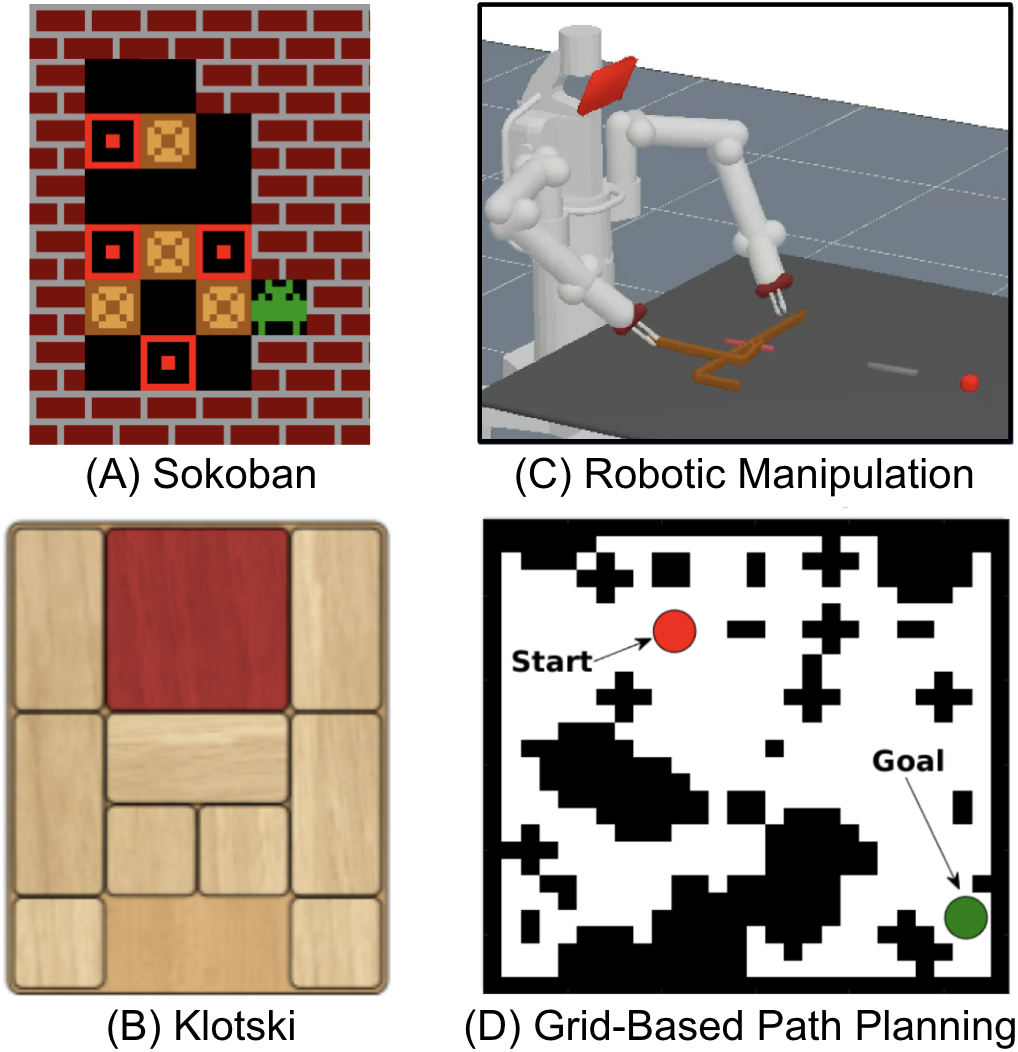}
\caption{Examples of environments related to \pw. (A) Sokoban 
\protect\cite{guez2019investigation}, (B) Klotski, (C) Robotic Manipulation 
\protect\cite{toussaint2018differentiable}, (D) Grid-Based Path Planning 
\protect\cite{TamarLA16}.}
\label{fig:related-envs-samples}
\end{figure}

Some robotic path planning tasks can be represented by discretizing the space of robot configurations into a grid (Fig.~\ref{fig:related-envs-samples}D) and then annotating which discrete configurations are collision-free \cite{TamarLA16,thrun2002probabilistic}. Planners must then solve for a collision-free path in the grid from a starting configuration to a goal configuration. Path planning is a sub-problem in \pw, so techniques for grid-based path planning, like hierarchical path planning \cite{harabor2008hierarchical}, are applicable.

\section{The Recursive Graph Distance Planner}

Existing classical planning algorithms \cite{fickert2018saarplan,helmert2006fast,seipp2018fast} are effective at logistics problems involving the constrained movements of multiple objects, but they are not optimized for planning motion paths. Simultaneously, existing algorithms for robotics \cite{dogar2011framework,garrett2020pddlstream,stilman2008planning} are designed for efficient path planning but are not typically applied in environments that require the diversity of obstacle and tool manipulations present in \pw puzzles. To reduce the gap between the specializations of existing algorithms, we introduce the \textbf{recursive graph distance} (RGD) heuristic, which optimizes a variation of the context-enhanced additive heuristic (CEA) \cite{helmert2008unifying} to achieve faster path planning. We then combine RGD with the novelty heuristic \cite{lipovetzky2017best} to explore efficiently in a greedy best-first (GBF) search.

\subsection{Planning Task Definition}

A multi-valued planning task $\Pi = (V, s_0, s_*, O)$ contains a set of variables $V$, an initial state $s_0$, a goal $s_*$, and a set of operators $O$ that map states to other states as a means to achieve the goal from the initial state.
A state $s$ is a mapping from variables $v \in V$ to values $d \in D_v$ in $v$'s domain $D_v$. For convenience we will write $(v, d) \in s$ to indicate that $s(v) = d$. A partial state is a mapping of a subset of variables in $V$ to values in their respective domains. In \pw, variables are typically object positions, although other state representations are possible. 
The goal $s_*$ is a partial state, and a state $s$ satisfies the goal whenever $s(v) = s_*(v)$ for all variables $v$ in the domain of $s_*$.

Operators can change one state into another, and each operator $o$ has a set of preconditions $\text{pre}(o)$ and a set of effects $\text{eff}(o)$, both of which are partial states. When a state $s$ satisfies an operator's preconditions, such that $s(v) = d$ for all $(v, d) \in \text{pre}(o)$, then applying the operator changes the state to $s'$ in which $s'(v) = d$ for all $(v, d) \in \text{eff}(o)$ and in which $s'(v) = s(v)$ for all other variables $v$ that are not in the domain of $\text{eff}(o)$.

\subsection{The Context-Enhanced Additive Heuristic}

The CEA heuristic estimates the number of operators needed to change state $s$ into a state in which the goal is satisfied. For each variable $v$, the heuristic uses a \textbf{domain transition graph} $\text{DTG}(v)$, which is a labeled directed multigraph with vertex set $D_v$ and an edge $(d, d')$ associated with each operator $o$ that has both $(v, d) \in \text{pre}(o)$ and $(v, d') \in \text{eff}(o)$. The graph may contain parallel edges if multiple operators can change the value of $v$ from $d$ to $d'$, and edges are labeled with the preconditions $\text{pre}(o)$ of their associated operators.

The CEA heuristic computes the sum of estimates of the number of operators needed to change the value $s(v)$ of each goal variable $v$ into a state $s'$ in which $s'(v) = s_*(v)$. Details of this computation are available in \cite{helmert2008unifying} and are not repeated here, but to estimate the cost of changing a variable $v$ from one value to a target value, the heuristic searches for a minimum-cost path in $\text{DTG}(v)$.

\begin{figure}[htbp]
\centering
\includegraphics[width=0.5\linewidth]{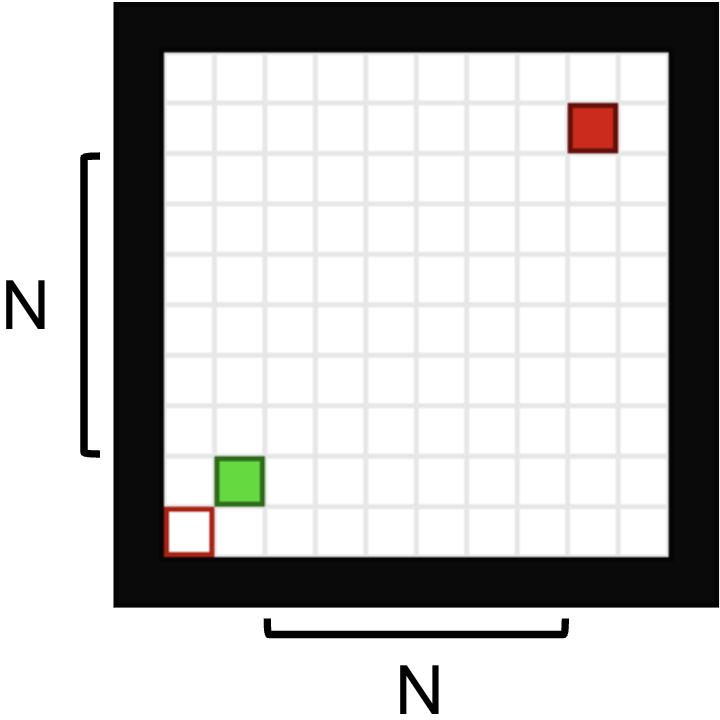}
\caption{The objects and the goal position remain in the corners as the puzzle scales with the distance $N$ between the objects. A greedy best-first search with the CEA heuristic \protect\cite{helmert2008unifying} solves this puzzle in $O(N^3)$ time, and the same search with the RGD heuristic solves this puzzle in $O(N^2)$ time.}
\label{fig:rgd_complex}
\end{figure}

If variables correspond to object positions in \pw, then in the puzzle shown in Fig~\ref{fig:rgd_complex}, computing the CEA heuristic for a single state has a worst-case complexity of $O(N^2)$ with respect to the distance $N$ between objects in the puzzle. This complexity results from the search for minimum-cost paths in DTGs, which each include $O(N^2)$ nodes for all object positions. A GBF search computes the heuristic in every explored state, resulting in $O(N^3)$ time to solve the puzzle.

\subsection{Recursive Graph Distance}

The RGD heuristic modifies the CEA heuristic to cache approximate costs of minimum-cost paths, and a GBF search with this heuristic requires $O(N^2)$ time to solve the puzzle in Fig~\ref{fig:rgd_complex}. Let $h^\text{RGD}(s)$ denote the estimated number of operators needed to achieve the goal $s_*$ starting from state $s$, and let $h_v^\text{RGD}(d|s)$ denote the estimated number of operators needed to change state $s$ into any state $s'$ in which $s'(v) = d$. The RGD heuristic estimates the cost to the goal by summing independent estimates of the cost to reach the goal value of each variable in the goal's domain:

\begin{equation}
h^\text{RGD}(s) = \sum_{v \in \text{dom}(s_*)}{h_v^\text{RGD}(s_*(v) |s)}
\end{equation}

Next, let $L_v(d, d')$ denote the number of edges in a minimum-length path from $d$ to $d'$ in $\text{DTG}(v)$, and let $\text{Succ}_v(d) = \{ ( d', p ) \}$ denote the set of end nodes $d'$ and corresponding labeled preconditions $p$ of all outgoing edges from $d$ in $\text{DTG}(v)$. To compute $h_v^\text{RGD}(d|s)$, we search over all operators that can modify the value of $v$ in $s$ to find which operator minimizes the sum of the remaining graph distance to $d$ after applying the operator, plus $1$ for the cost of the operator, plus the estimated cost of the operator's preconditions:

\begin{equation}
  \begin{split}
    h_v^\text{RGD}(d|s) = \min_{( d', p ) \in \text{Succ}_v(s(v))} \biggl[L_v&(d',d) + 1 +\\
    &\sum_{(v_i, e_i) \in p}{h^\text{RGD}_{v_i}(e_i|s)} \biggr]
    \end{split}
\end{equation}

\begin{figure}[htbp]
\centering
\includegraphics[width=0.6\linewidth]{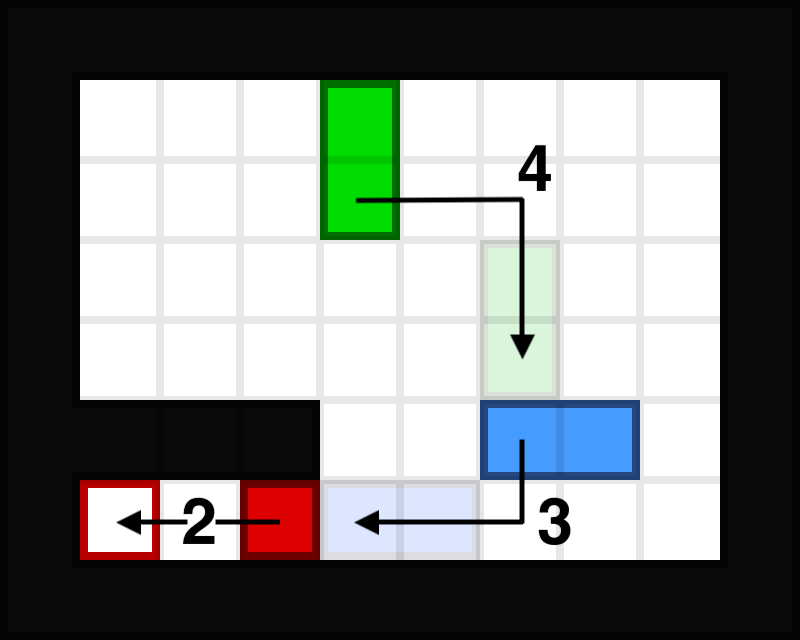}
\caption{An illustration of the RGD heuristic. The red object must move two positions left to reach its goal position, but the actor (green) is unable to push it due to the walls. RGD detects that the blue object can push the red object, and it recursively discovers that the actor can move into a position to push the blue object along a path toward the red object. RGD finds the minimum-length paths to move objects into either a pushing position or a goal position. The actual number of actions required to achieve the goal is 12, but RGD estimates 4+3+2 = 9 actions because it only considers the cost of pushing the blue object by one position; it ignores the cost of moving the actor to continue pushing the blue object along its path. To estimate the minimum number of actions to achieve a goal, RGD explores all pushing directions, all possible pushers, and all relative pushing positions between objects.}
\label{fig:rgd}
\end{figure}

We also apply two exceptions: whenever $s(v) = d$, we return $0$ from $h_v^\text{RGD}(d|s)$, and to avoid infinite recursion, we return infinity whenever $v$ appears in any recursive parent call of $h_v^\text{RGD}(d|s)$. Figure~\ref{fig:rgd} illustrates calculating the RGD heuristic in \pw. Domain transition graphs are computed once for all variables at the beginning of each search that uses this heuristic.

This formulation allows $h^\text{RGD}(s)$ to cache the shortest path lengths $L_v(d, d')$. Continuing with the example in Fig~\ref{fig:rgd_complex}, the minimum-length path to move the agent to the red object requires $O(N^2)$ time to compute in the DTG for the agent's position, and likewise for the path from the red object's position to its goal. In a GBF search, expanded states can reuse the minimum-length paths from those computed while expanding preceding states. Consequently, in this example the heuristic requires $O(N^2)$ time when it is first computed, and in all subsequent expanded states it requires $O(1)$ time, resulting in a total time of $O(N^2)$ to solve the puzzle in a GBF search.

Appendix \ref{appendix.rgd} provides pseudocode for RGD and includes additional optimizations. We also provide an open-source C++ implementation.\footnote[2]{\pwcodeurl}

\subsection{Combining with the Novelty Heuristic}

The novelty heuristic \cite{lipovetzky2017best} promotes exploration from states that contain previously unseen partial states. For some set of generated states $S = \{s_i\}$, the \textbf{novelty} $w(s)$ of a newly generated state $s$ is the size of the smallest non-empty set of variables $V$ such that there is no state $s_i \in S$ for which $s(v) = s_i(v)$ for all $v \in V$. In other words, $V$ defines a set of variables whose combined values occur in $s$ but have not yet occurred in any other generated state, and $w(s)$ measures the size of the smallest such set.

When variables correspond to the positions of objects in a \pw puzzle, then $w(s) = 1$ if $s$ is the first state in which some object has a position that has not occurred in any previously generated state. If $s$ is the first state in which some pair of objects have positions that have not jointly occurred in any previously generated state, then $w(s) = 2$. 

We define the \textbf{Novelty+RGD planner} as a GBF search that prioritizes states using a lexicographic combination of heuristics: first prioritize states that minimize $w(s)$, and whenever states have the same novelty, then prioritize states that minimize $h^\text{RGD}(s)$. In \pw this planner defines a state $s$ as a mapping from object identifiers to their 2D positions, and because computing novelty is combinatorially expensive in the number of objects, the planner limits the maximum novelty to 3 to reduce the novelty computation to polynomial time.

\section{Evaluation}

\subsection{Benchmarking Dataset}

To compare how efficiently different algorithms can solve \pw puzzles, we provide a set of 223 hand-designed puzzles, which are available to play online and download.\footnote[3]{\pwplayurl} Puzzles are organized into four levels of difficulty, such that people can typically solve Level 1 puzzles within seconds, Level 2 puzzles within a few minutes, Level 3 puzzles within 15 minutes, and longer for Level 4. From levels 1 to 4, the dataset contains 68, 74, 67, and 14 puzzles, respectively.

\subsection{Classical Planning}

To evaluate the performance of existing classical planners, we aimed to cover multiple classes of search heuristics, including delete-relaxations \cite{hoffmann2001ff}, causal graphs \cite{helmert2006fast}, landmarks \cite{richter2010lama}, and state novelty \cite{frances2018best}, as well as portfolios containing multiple planners \cite{fickert2018saarplan,seipp2018fast}. We preferred the winning planners from International Planning Competitions (IPC), resulting in the following selection:

\begin{itemize}
  \item \textbf{Fast Forward} (FF) \cite{hoffmann2001ff} is a seminal planning algorithm that relies on a delete relaxation heuristic.

  \item \textbf{Fast Downward} (FD) \cite{helmert2006fast} introduced DTGs and the causal graph heuristic (winner of IPC 2004 classical track). We selected the most recent implementation, which uses the CEA heuristic.

  \item \textbf{LAMA} \cite{richter2010lama} introduced the landmark heuristic (winner of IPC 2008 sequential satisficing track).

  \item \textbf{Best-First Width Search} (BFWS) \cite{frances2018best} uses the novelty heuristic, which prioritizes states that contain novel substates (winner of IPC 2018 agile track).
    
  \item \textbf{Fast Downward Stone Soup} (FDSS) \cite{seipp2018fast} is a portfolio of planners (winner of IPC 2018 satisficing track).
    
  \item \textbf{Saarplan} \cite{fickert2018saarplan} is another portfolio of planners (runner-up of IPC 2018 agile track).
\end{itemize}

Multiple versions of these planners are available, and Appendix \ref{appendix.classical_planners} explains the versions we selected. In particular, we found that among the available variations of BFWS, Dual BFWS performed best in \pw. All planners are implemented in C or C++.

Some planners are optimized for different problem representations, like SAS+ \cite{backstrom1995complexity} or PDDL \cite{ghallab98pddl}. Translating between representations can exceed the search time in \pw puzzles. Because these representations make no assumptions that are specific to \pw's dynamics, we omitted translation times from the reported planning times to avoid penalizing planners for their problem representations.

\subsection{Model-Free Reinforcement Learning}

We selected two seminal, diverse deep RL algorithms to evaluate on \pw: \textbf{Deep Q-Network (DQN)} \cite{mnih2015human}, an off-policy, value-based algorithm, and \textbf{Proximal Policy Optimization (PPO)} \cite{schulman2017proximal}, an on-policy, policy-gradient algorithm. We chose these algorithms because they are widely used, easy to implement, and have shown competitive performance in environments like Atari games and continuous robotic control. Furthermore, recent work \cite{guez2019investigation} has shown that model-free approaches can be successful at solving tasks that may seem more suited for model-based approaches. Appendix \ref{appendix.network_architecture} provides details on the network architectures and hyperparameter settings we used in experiments.

We implemented \pw as an OpenAI Gym environment \cite{gym} with a reward function similar to that in \cite{guez2019investigation}: the agent receives +1 reward for moving an object onto its goal position and -1 for moving off of its goal position. The agent receives +10 reward for achieving all goals in a puzzle. In contrast to \cite{guez2019investigation}, the agent receives -0.01 reward for each action, since solutions to \pw puzzles tend to be longer than solutions to Sokobans.

DQN and PPO require more training data than the 223 hand-crafted puzzles can provide, so we procedurally generated sets of puzzles to provide sufficient training data. To measure how puzzle parameters affect RL algorithms, we generated the following puzzle sets:

\begin{itemize}
    \item \textbf{Base:} The puzzle grid size is 5x5 discrete positions. Each puzzle contains three walls, one object with a goal position, and one movable obstacle. All objects have 1x1 shape, including the walls and the agent. All objects and goal positions are placed randomly.
    \item \textbf{Larger Puzzle Sizes:} Same as Base, but the puzzle width and height are independently uniformly sampled between 5 and 10, inclusive.
    \item \textbf{More Walls:} Same as Base, but the number of walls is uniformly sampled between 3 and 5, inclusive.
    \item \textbf{More Obstacles:} Same as Base, but with two movable obstacle objects.
    \item \textbf{More Shapes:} Same as Base, but the shape of the agent, obstacle, and target object are independently uniformly sampled polyominoes with 1, 2, or 3 squares.
    \item \textbf{Multiple Goals:} Same as Base, but with two objects with goal positions. To distinguish these objects, one has shape 1x1 and the other 1x2.
    \item \textbf{All:} All previous variations combined simultaneously. Each puzzle's width and height are uniformly sampled between 5 and 10, the number of walls is uniformly sampled between 3 and 5, the number of movable obstacles is uniformly sampled between 1 and 2, the number of goal objects is uniformly sampled between 1 and 2, and object shapes are uniformly sampled polyominoes with between 1 and 3 squares.
\end{itemize}

These puzzles are limited to small sizes to mitigate sparse reward; an RL agent must initially be able to solve some puzzles by chance so that it can learn how to achieve goals. Because these puzzle sets are generally easier than the hand-designed Level 1 puzzles, we designate them as Level 0.

For each Level 0 variation, we generated a set of 2,000 puzzles for training and 200 more for testing. This puzzle generating process employed a classical planner to guarantee that all generated puzzles were solvable. To reduce bias during both training and testing, we augmented the puzzles by including every combination of 90 degree rotation and flipping, resulting in 16,000 augmented puzzles per set. For variations involving puzzle size changes, we added walls around all puzzles to pad them up to the largest puzzle's size. The position of each smaller puzzle was randomized within the padded walls, providing additional augmentation.

To evaluate performance on the hand-designed puzzles, we trained each algorithm on all puzzles in the Level 1 set and included the same augmentations of rotation, flipping, and padding puzzles up to the largest size. For this evaluation we did not pretrain on the Level 0 puzzle sets. To give the algorithms the highest chance of learning from the training puzzles, we did not use a train/test split because the results showed that few training puzzles were solved. Due to their low performance on Level 1, we did not evaluate PPO or DQN on Levels 2 -- 4.

\section{Experimental Results}

\begin{figure}
\centering
\includegraphics[width=\linewidth]{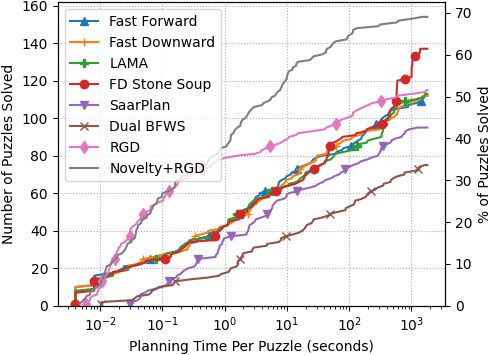}
\caption{Number of solved puzzles vs. planning time. Planning time is measured independently for each puzzle.}
\label{fig:results-planner-all}
\end{figure}

\subsection{Classical Planning Results}

We evaluated all classical planners on the set of hand-designed puzzles. All planners ran on a 2.25 GHz AMD CPU, subject to a 30 GB memory limit and a 30 minute time limit per puzzle. Results are shown in Figure~\ref{fig:results-planner-all}. Among existing classical planners, FDSS performed the best, solving 61.4\% of the puzzles. Despite its relative simplicity, Fast Forward was competitive with most other planners. The RGD plot shows the performance of a greedy best-first search using the RGD heuristic, which outperformed all existing classical planners except FDSS. The Novelty+RGD planner performed best, solving 69.1\% of the puzzles. In fact, Novelty+RGD solved as many puzzles in 45 seconds as FDSS solved in 30 minutes, which amounts to a 40x speed improvement.

\subsection{Deep Reinforcement Learning Results}

\subsubsection{PPO and DQN on Level 0}

Table~\ref{ppo_dqn_level0_results} shows PPO and DQN results on Level 0. 
PPO's training accuracy is significantly higher than its test accuracy, indicating possible overfitting. We found that this phenomenon persisted despite tuning the network capacity. By contrast, DQN shows a smaller discrepancy between training and testing, suggesting that DQN's learned value function is more transferable across puzzles than PPO's learned policy.

Both PPO and DQN perform best on \textbf{Larger Puzzle Sizes}. We believe this is because these puzzles have the lowest spatial density of walls and movable obstacles. Both algorithms perform worst on \textbf{Multiple Goals}. We suspect this is because puzzles with multiple goals are less likely to solve by chance and because some puzzles become unsolvable if goals are achieved in the wrong order.
Both algorithms also have low performance on \textbf{All}, indicating that the diversity in puzzles hinders both training and transfer to unseen puzzles from the same distribution.

\begin{table}
  \centering
  \begin{tabular}{l|cc|cc}
  {} & \multicolumn{2}{c}{PPO} & \multicolumn{2}{c}{DQN}\\
  Level 0 Puzzle Set & Train & Test & Train & Test \\
  \hline
  Base & 88.5\% & 40.4\% & 24.9\% & 19.5\% \\
  Larger Puzzle Sizes & 93.2\% & 71.5\% & 61.2\% & 61.8\% \\
  More Walls & 77.9\% & 23.3\% & 18.5\% & 13.1\% \\
  More Obstacles & 64.7\% & 21.4\% & 17.6\% & 13.1\% \\
  More Shapes & 74.5\% & 24.2\% & 23.9\% & 17.1\% \\
  Multiple Goals & 5.3\% & 1.1\% & 0.9\% & 0.2\% \\
  All & 21.2\% & 5.7\% & 11.4\% & 10.1\% \\
  \end{tabular}
  \caption{The percentage of training and testing puzzles solved by PPO and DQN.
  }
  \label{ppo_dqn_level0_results}
\end{table}

\subsubsection{PPO and DQN on Level 1}

Compared to the Level 0 puzzles, Level 1 puzzles are generally larger, include more object shapes, and involve more complex arrangements of obstacles, walls, and goals. Without pretraining on Level 0 puzzle sets, we trained both PPO and DQN on all Level 1 puzzles for 350 million steps and measured their training performance, the percentage of Level 1 puzzles they could solve. DQN converged to solving less than 1\% of the puzzles, and PPO converged to solving 6\% of the puzzles. We believe this low performance is in part due to the low probability of solving most Level 1 puzzles with the initial policy, resulting in sparse positive rewards.

\section{Conclusions \& Discussion}

We present \pw, a manipulation planning environment involving tools and movable obstacles. Compared to similar environments such as Sokoban, we believe \pw presents many novel challenges relevant for research communities such as classical planning, reinforcement learning, and robotics. In our baseline evaluations we introduced the RGD+Novelty  planner, which outperformed many state-of-the-art classical planners and RL algorithms on \pw, yet is still below human-level performance. To close this gap, we hope \pw can become a standard benchmark to drive forward research in RL and physical planning.

For future work, we believe in exploring the following directions: First, hierarchical planning \cite{kaelbling2010hierarchical} and factorized task and motion planning \cite{garrett2020pddlstream} may be effective for \pw. Second, model-based RL algorithms such as AlphaZero \cite{silver2017mastering} and MuZero \cite{schrittwieser2020mastering} can potentially discover new and interesting heuristics. Third, many RL research areas could improve both training efficiency and generalization to unseen \pw puzzles: hierarchical RL, credit assignment \cite{cca} using a causal model \cite{kansky2017schema}, efficient exploration, multi-task learning \cite{https://doi.org/10.48550/arxiv.1611.01796}, meta-learning \cite{https://doi.org/10.48550/arxiv.1703.03400}, and curriculum learning \cite{narvekar2020curriculum}. Finally, a cognitive science-inspired analysis of how people solve \pw puzzles may help develop better algorithms.

The \pw environment and benchmark puzzles are open source, and we encourage the community to expand on this benchmark. We hope that lessons learned in \pw will translate into more intelligent systems in real-world applications.

\bibliographystyle{named}
\bibliography{references}

\begin{thebibliography}{}

\bibitem[\protect\citeauthoryear{Andreas \bgroup \em et al.\egroup
  }{2016}]{https://doi.org/10.48550/arxiv.1611.01796}
Jacob Andreas, Dan Klein, and Sergey Levine.
\newblock Modular multitask reinforcement learning with policy sketches, 2016.

\bibitem[\protect\citeauthoryear{Andrychowicz \bgroup \em et al.\egroup
  }{2017}]{her}
Marcin Andrychowicz, Filip Wolski, Alex Ray, Jonas Schneider, Rachel Fong,
  Peter Welinder, Bob McGrew, Josh Tobin, Pieter Abbeel, and Wojciech Zaremba.
\newblock Hindsight experience replay, 2017.

\bibitem[\protect\citeauthoryear{B{\"a}ckstr{\"o}m and
  Nebel}{1995}]{backstrom1995complexity}
Christer B{\"a}ckstr{\"o}m and Bernhard Nebel.
\newblock Complexity results for sas+ planning.
\newblock {\em Computational Intelligence}, 11(4):625--655, 1995.

\bibitem[\protect\citeauthoryear{Bellemare \bgroup \em et al.\egroup
  }{2016}]{https://doi.org/10.48550/arxiv.1606.01868}
Marc~G. Bellemare, Sriram Srinivasan, Georg Ostrovski, Tom Schaul, David
  Saxton, and Remi Munos.
\newblock Unifying count-based exploration and intrinsic motivation, 2016.

\bibitem[\protect\citeauthoryear{Brockman \bgroup \em et al.\egroup
  }{2016}]{gym}
Greg Brockman, Vicki Cheung, Ludwig Pettersson, Jonas Schneider, John Schulman,
  Jie Tang, and Wojciech Zaremba.
\newblock Openai gym, 2016.

\bibitem[\protect\citeauthoryear{Dogar and
  Srinivasa}{2011}]{dogar2011framework}
Mehmet Dogar and Siddhartha Srinivasa.
\newblock A framework for push-grasping in clutter.
\newblock {\em Robotics: Science and systems VII}, 1, 2011.

\bibitem[\protect\citeauthoryear{Dor and Zwick}{1999}]{dor1999sokoban}
Dorit Dor and Uri Zwick.
\newblock Sokoban and other motion planning problems.
\newblock {\em Computational Geometry}, 13(4):215--228, 1999.

\bibitem[\protect\citeauthoryear{Fickert \bgroup \em et al.\egroup
  }{2018}]{fickert2018saarplan}
Maximilian Fickert, Daniel Gnad, Patrick Speicher, and J{\"o}rg Hoffmann.
\newblock Saarplan: Combining saarland’s greatest planning techniques.
\newblock {\em IPC2018--Classical Tracks}, pages 10--15, 2018.

\bibitem[\protect\citeauthoryear{Finn \bgroup \em et al.\egroup
  }{2017}]{https://doi.org/10.48550/arxiv.1703.03400}
Chelsea Finn, Pieter Abbeel, and Sergey Levine.
\newblock Model-agnostic meta-learning for fast adaptation of deep networks,
  2017.

\bibitem[\protect\citeauthoryear{Florensa \bgroup \em et al.\egroup
  }{2017}]{florensa2017stochastic}
Carlos Florensa, Yan Duan, and Pieter Abbeel.
\newblock Stochastic neural networks for hierarchical reinforcement learning.
\newblock {\em arXiv preprint arXiv:1704.03012}, 2017.

\bibitem[\protect\citeauthoryear{Frances \bgroup \em et al.\egroup
  }{2018}]{frances2018best}
Guillem Frances, Hector Geffner, Nir Lipovetzky, and Miquel Ramirez.
\newblock Best-first width search in the ipc 2018: Complete, simulated, and
  polynomial variants.
\newblock {\em IPC2018--Classical Tracks}, pages 22--26, 2018.

\bibitem[\protect\citeauthoryear{Garg \bgroup \em et al.\egroup
  }{2022}]{https://doi.org/10.48550/arxiv.2203.00054}
Divyansh Garg, Skanda Vaidyanath, Kuno Kim, Jiaming Song, and Stefano Ermon.
\newblock Lisa: Learning interpretable skill abstractions from language, 2022.

\bibitem[\protect\citeauthoryear{Garrett \bgroup \em et al.\egroup
  }{2020}]{garrett2020pddlstream}
Caelan~Reed Garrett, Tom{\'a}s Lozano-P{\'e}rez, and Leslie~Pack Kaelbling.
\newblock Pddlstream: Integrating symbolic planners and blackbox samplers via
  optimistic adaptive planning.
\newblock In {\em Proceedings of the International Conference on Automated
  Planning and Scheduling}, volume~30, pages 440--448, 2020.

\bibitem[\protect\citeauthoryear{Ghallab \bgroup \em et al.\egroup
  }{1998}]{ghallab98pddl}
M.~Ghallab, A.~Howe, C.~Knoblock, D.~Mcdermott, A.~Ram, M.~Veloso, D.~Weld, and
  D.~Wilkins.
\newblock {PDDL - The Planning Domain Definition Language}.
\newblock {\em Technical Report, Tech. Rep.}, 1998.

\bibitem[\protect\citeauthoryear{Grinsztajn \bgroup \em et al.\egroup
  }{2021}]{https://doi.org/10.48550/arxiv.2106.04480}
Nathan Grinsztajn, Johan Ferret, Olivier Pietquin, Philippe Preux, and Matthieu
  Geist.
\newblock There is no turning back: A self-supervised approach for
  reversibility-aware reinforcement learning, 2021.

\bibitem[\protect\citeauthoryear{Guez \bgroup \em et al.\egroup
  }{2019}]{guez2019investigation}
Arthur Guez, Mehdi Mirza, Karol Gregor, Rishabh Kabra, S{\'e}bastien
  Racani{\`e}re, Th{\'e}ophane Weber, David Raposo, Adam Santoro, Laurent
  Orseau, et~al.
\newblock An investigation of model-free planning.
\newblock In {\em International Conference on Machine Learning}, pages
  2464--2473. PMLR, 2019.

\bibitem[\protect\citeauthoryear{Hamrick \bgroup \em et al.\egroup
  }{2020}]{hamrick2020role}
Jessica~B Hamrick, Abram~L Friesen, Feryal Behbahani, Arthur Guez, Fabio Viola,
  Sims Witherspoon, Thomas Anthony, Lars Buesing, Petar Veli{\v{c}}kovi{\'c},
  and Th{\'e}ophane Weber.
\newblock On the role of planning in model-based deep reinforcement learning.
\newblock {\em arXiv preprint arXiv:2011.04021}, 2020.

\bibitem[\protect\citeauthoryear{Harabor and
  Botea}{2008}]{harabor2008hierarchical}
Daniel Harabor and Adi Botea.
\newblock Hierarchical path planning for multi-size agents in heterogeneous
  environments.
\newblock In {\em 2008 IEEE Symposium On Computational Intelligence and Games},
  pages 258--265. IEEE, 2008.

\bibitem[\protect\citeauthoryear{Harutyunyan \bgroup \em et al.\egroup
  }{2019}]{hca}
Anna Harutyunyan, Will Dabney, Thomas Mesnard, Mohammad Gheshlaghi~Azar, Bilal
  Piot, Nicolas Heess, Hado~P van Hasselt, Gregory Wayne, Satinder Singh, Doina
  Precup, et~al.
\newblock Hindsight credit assignment.
\newblock {\em Advances in neural information processing systems}, 32, 2019.

\bibitem[\protect\citeauthoryear{Haslum \bgroup \em et al.\egroup
  }{2007}]{haslum2007domain}
Patrik Haslum, Adi Botea, Malte Helmert, Blai Bonet, Sven Koenig, et~al.
\newblock Domain-independent construction of pattern database heuristics for
  cost-optimal planning.
\newblock In {\em AAAI}, volume~7, pages 1007--1012, 2007.

\bibitem[\protect\citeauthoryear{Helmert and
  Geffner}{2008}]{helmert2008unifying}
Malte Helmert and H{\'e}ctor Geffner.
\newblock Unifying the causal graph and additive heuristics.
\newblock In {\em ICAPS}, pages 140--147, 2008.

\bibitem[\protect\citeauthoryear{Helmert}{2006}]{helmert2006fast}
Malte Helmert.
\newblock The fast downward planning system.
\newblock {\em Journal of Artificial Intelligence Research}, 26:191--246, 2006.

\bibitem[\protect\citeauthoryear{Hoffman \bgroup \em et al.\egroup
  }{2020}]{hoffman2020acme}
Matt Hoffman, Bobak Shahriari, John Aslanides, Gabriel Barth-Maron, Feryal
  Behbahani, Tamara Norman, Abbas Abdolmaleki, Albin Cassirer, Fan Yang, Kate
  Baumli, et~al.
\newblock Acme: A research framework for distributed reinforcement learning.
\newblock {\em arXiv preprint arXiv:2006.00979}, 2020.

\bibitem[\protect\citeauthoryear{Hoffmann and Nebel}{2001}]{hoffmann2001ff}
J{\"o}rg Hoffmann and Bernhard Nebel.
\newblock The ff planning system: Fast plan generation through heuristic
  search.
\newblock {\em Journal of Artificial Intelligence Research}, 14:253--302, 2001.

\bibitem[\protect\citeauthoryear{Hong \bgroup \em et al.\egroup
  }{2018}]{NEURIPS2018_a2802cad}
Zhang-Wei Hong, Tzu-Yun Shann, Shih-Yang Su, Yi-Hsiang Chang, Tsu-Jui Fu, and
  Chun-Yi Lee.
\newblock Diversity-driven exploration strategy for deep reinforcement
  learning.
\newblock In S.~Bengio, H.~Wallach, H.~Larochelle, K.~Grauman, N.~Cesa-Bianchi,
  and R.~Garnett, editors, {\em Advances in Neural Information Processing
  Systems}, volume~31. Curran Associates, Inc., 2018.

\bibitem[\protect\citeauthoryear{Kaelbling and
  Lozano-P{\'e}rez}{2010}]{kaelbling2010hierarchical}
Leslie~Pack Kaelbling and Tom{\'a}s Lozano-P{\'e}rez.
\newblock Hierarchical planning in the now.
\newblock In {\em Workshops at the Twenty-Fourth AAAI Conference on Artificial
  Intelligence}, 2010.

\bibitem[\protect\citeauthoryear{Kansky \bgroup \em et al.\egroup
  }{2017}]{kansky2017schema}
Ken Kansky, Tom Silver, David~A M{\'e}ly, Mohamed Eldawy, Miguel
  L{\'a}zaro-Gredilla, Xinghua Lou, Nimrod Dorfman, Szymon Sidor, Scott
  Phoenix, and Dileep George.
\newblock Schema networks: Zero-shot transfer with a generative causal model of
  intuitive physics.
\newblock In {\em International conference on machine learning}, pages
  1809--1818. PMLR, 2017.

\bibitem[\protect\citeauthoryear{Lecun \bgroup \em et al.\egroup
  }{1998}]{726791}
Y.~Lecun, L.~Bottou, Y.~Bengio, and P.~Haffner.
\newblock Gradient-based learning applied to document recognition.
\newblock {\em Proceedings of the IEEE}, 86(11):2278--2324, 1998.

\bibitem[\protect\citeauthoryear{Li \bgroup \em et al.\egroup
  }{2019}]{https://doi.org/10.48550/arxiv.1906.05862}
Alexander~C. Li, Carlos Florensa, Ignasi Clavera, and Pieter Abbeel.
\newblock Sub-policy adaptation for hierarchical reinforcement learning, 2019.

\bibitem[\protect\citeauthoryear{Lipovetzky and
  Geffner}{2017}]{lipovetzky2017best}
Nir Lipovetzky and Hector Geffner.
\newblock Best-first width search: Exploration and exploitation in classical
  planning.
\newblock In {\em Thirty-First AAAI Conference on Artificial Intelligence},
  2017.

\bibitem[\protect\citeauthoryear{Liu \bgroup \em et al.\egroup
  }{2020}]{https://doi.org/10.48550/arxiv.2008.02790}
Evan~Zheran Liu, Aditi Raghunathan, Percy Liang, and Chelsea Finn.
\newblock Decoupling exploration and exploitation for meta-reinforcement
  learning without sacrifices, 2020.

\bibitem[\protect\citeauthoryear{Mesnard \bgroup \em et al.\egroup
  }{2020}]{cca}
Thomas Mesnard, Théophane Weber, Fabio Viola, Shantanu Thakoor, Alaa Saade,
  Anna Harutyunyan, Will Dabney, Tom Stepleton, Nicolas Heess, Arthur Guez,
  Éric Moulines, Marcus Hutter, Lars Buesing, and Rémi Munos.
\newblock Counterfactual credit assignment in model-free reinforcement
  learning, 2020.

\bibitem[\protect\citeauthoryear{Mnih \bgroup \em et al.\egroup
  }{2015}]{mnih2015human}
Volodymyr Mnih, Koray Kavukcuoglu, David Silver, Andrei~A Rusu, Joel Veness,
  Marc~G Bellemare, Alex Graves, Martin Riedmiller, Andreas~K Fidjeland, Georg
  Ostrovski, et~al.
\newblock Human-level control through deep reinforcement learning.
\newblock {\em nature}, 518(7540):529--533, 2015.

\bibitem[\protect\citeauthoryear{Nachum \bgroup \em et al.\egroup
  }{2018}]{https://doi.org/10.48550/arxiv.1805.08296}
Ofir Nachum, Shixiang Gu, Honglak Lee, and Sergey Levine.
\newblock Data-efficient hierarchical reinforcement learning, 2018.

\bibitem[\protect\citeauthoryear{Narvekar \bgroup \em et al.\egroup
  }{2020}]{narvekar2020curriculum}
Sanmit Narvekar, Bei Peng, Matteo Leonetti, Jivko Sinapov, Matthew~E Taylor,
  and Peter Stone.
\newblock Curriculum learning for reinforcement learning domains: A framework
  and survey.
\newblock {\em arXiv preprint}, 2020.

\bibitem[\protect\citeauthoryear{Richter and Westphal}{2010}]{richter2010lama}
Silvia Richter and Matthias Westphal.
\newblock The lama planner: Guiding cost-based anytime planning with landmarks.
\newblock {\em Journal of Artificial Intelligence Research}, 39:127--177, 2010.

\bibitem[\protect\citeauthoryear{Schrittwieser \bgroup \em et al.\egroup
  }{2020}]{schrittwieser2020mastering}
Julian Schrittwieser, Ioannis Antonoglou, Thomas Hubert, Karen Simonyan,
  Laurent Sifre, Simon Schmitt, Arthur Guez, Edward Lockhart, Demis Hassabis,
  Thore Graepel, et~al.
\newblock Mastering atari, go, chess and shogi by planning with a learned
  model.
\newblock {\em Nature}, 588(7839):604--609, 2020.

\bibitem[\protect\citeauthoryear{Schulman \bgroup \em et al.\egroup
  }{2017}]{schulman2017proximal}
John Schulman, Filip Wolski, Prafulla Dhariwal, Alec Radford, and Oleg Klimov.
\newblock Proximal policy optimization algorithms.
\newblock {\em arXiv preprint arXiv:1707.06347}, 2017.

\bibitem[\protect\citeauthoryear{Seipp and R{\"o}ger}{2018}]{seipp2018fast}
Jendrik Seipp and Gabriele R{\"o}ger.
\newblock Fast downward stone soup 2018.
\newblock {\em IPC2018--Classical Tracks}, pages 72--74, 2018.

\bibitem[\protect\citeauthoryear{Shoham and Schaeffer}{2020}]{shoham2020fess}
Yaron Shoham and Jonathan Schaeffer.
\newblock The fess algorithm: A feature based approach to single-agent search.
\newblock In {\em 2020 IEEE Conference on Games (CoG)}, pages 96--103. IEEE,
  2020.

\bibitem[\protect\citeauthoryear{Silver \bgroup \em et al.\egroup
  }{2017}]{silver2017mastering}
David Silver, Thomas Hubert, Julian Schrittwieser, Ioannis Antonoglou, Matthew
  Lai, Arthur Guez, Marc Lanctot, Laurent Sifre, Dharshan Kumaran, Thore
  Graepel, et~al.
\newblock Mastering chess and shogi by self-play with a general reinforcement
  learning algorithm.
\newblock {\em arXiv preprint arXiv:1712.01815}, 2017.

\bibitem[\protect\citeauthoryear{Sim{\'e}on \bgroup \em et al.\egroup
  }{2004}]{simeon2004manipulation}
Thierry Sim{\'e}on, Jean-Paul Laumond, Juan Cort{\'e}s, and Anis Sahbani.
\newblock Manipulation planning with probabilistic roadmaps.
\newblock {\em The International Journal of Robotics Research},
  23(7-8):729--746, 2004.

\bibitem[\protect\citeauthoryear{Stilman and
  Kuffner}{2008}]{stilman2008planning}
Mike Stilman and James Kuffner.
\newblock Planning among movable obstacles with artificial constraints.
\newblock {\em The International Journal of Robotics Research},
  27(11-12):1295--1307, 2008.

\bibitem[\protect\citeauthoryear{Stilman \bgroup \em et al.\egroup
  }{2007}]{stilman2007manipulation}
Mike Stilman, Jan-Ullrich Schamburek, James Kuffner, and Tamim Asfour.
\newblock Manipulation planning among movable obstacles.
\newblock In {\em Proceedings 2007 IEEE international conference on robotics
  and automation}, pages 3327--3332. IEEE, 2007.

\bibitem[\protect\citeauthoryear{Tamar \bgroup \em et al.\egroup
  }{2016}]{TamarLA16}
Aviv Tamar, Sergey Levine, and Pieter Abbeel.
\newblock Value iteration networks.
\newblock {\em CoRR}, 2016.

\bibitem[\protect\citeauthoryear{Team \bgroup \em et al.\egroup
  }{2021}]{team2021open}
Open Ended~Learning Team, Adam Stooke, Anuj Mahajan, Catarina Barros, Charlie
  Deck, Jakob Bauer, Jakub Sygnowski, Maja Trebacz, et~al.
\newblock Open-ended learning leads to generally capable agents.
\newblock {\em arXiv preprint arXiv:2107.12808}, 2021.

\bibitem[\protect\citeauthoryear{Thrun}{2002}]{thrun2002probabilistic}
Sebastian Thrun.
\newblock Probabilistic robotics.
\newblock {\em Communications of the ACM}, 45(3):52--57, 2002.

\bibitem[\protect\citeauthoryear{Todorov \bgroup \em et al.\egroup
  }{2012}]{todorov2012mujoco}
Emanuel Todorov, Tom Erez, and Yuval Tassa.
\newblock Mujoco: A physics engine for model-based control.
\newblock In {\em 2012 IEEE/RSJ International Conference on Intelligent Robots
  and Systems}, pages 5026--5033. IEEE, 2012.

\bibitem[\protect\citeauthoryear{Toussaint \bgroup \em et al.\egroup
  }{2018}]{toussaint2018differentiable}
Marc~A Toussaint, Kelsey~Rebecca Allen, Kevin~A Smith, and Joshua~B Tenenbaum.
\newblock Differentiable physics and stable modes for tool-use and manipulation
  planning.
\newblock {\em Robotics: Science and Systems 2018}, 2018.

\bibitem[\protect\citeauthoryear{Yan \bgroup \em et al.\egroup
  }{2022}]{yan2022deep}
Dong Yan, Jiayi Weng, Shiyu Huang, Chongxuan Li, Yichi Zhou, Hang Su, and Jun
  Zhu.
\newblock Deep reinforcement learning with credit assignment for combinatorial
  optimization.
\newblock {\em Pattern Recognition}, 124:108466, 2022.

\end{thebibliography}

\appendix

\section{Appendix}

\subsection{\pw Puzzle Definition}
\label{appendix.puzzle_def}

Formally, a \pw puzzle is a 4-tuple $(S, A, G, T)$ with the following components:

$S$ is a finite set of states, where each state $s$ is a set of object states.
An \textbf{object state} $o_i$ is a pair $(p_i, Y_i)$ containing the 2D integer position $p_i$ of the object and a set $Y_i$ of the 2D integer positions of all cells that the object occupies relative to $p_i$. One object $o_A$ is designated as the agent, which is directly controlled by actions, and another object $o_W$ is designated as an immovable wall. For convenience, let $Z_i(p) = \{ y + p \mid y \in Y_i \}$ denote the set of absolute positions of all cells in object $i$ when the object is located at position $p$. No object's occupied cells are allowed to intersect the occupied cells of any other object in any state: $Z_i(p_i) \cap Z_j(p_j) = \emptyset$ for all object pairs $i$, $j$.

$A$ is a fixed set of actions $\{\textrm{Left}, \textrm{Right}, \textrm{Up}, \textrm{Down}\}$, which correspond to moving the agent by a single cell in the corresponding direction. For each action $a$, let $\hat{u}_a$ denote the unit vector in the direction of movement.

$G$ defines the goal of the puzzle. It is a mapping from one or more object indices $i$ to corresponding goal positions $g_i$. The goal of the puzzle is achieved in a given state when $p_i = g_i$ for all $(i, g_i)$ pairs in $G$.

$T$ is a state transition function that maps $S \times A \rightarrow S$. Let $s' = T(s, a)$, and likewise let $p_i'$ denote the position of object $i$ in $s'$. For every action $a$, a subset of objects $M$ may move in the direction $\hat{u}_a$. $M$ contains the agent object as well as every object $m$ for which there exists a chain of two or more objects ${c_1, c_2, ..., c_n}$ that satisfy these conditions:

\begin{itemize}
    \item $c_1$ is the agent, which is the root cause of the movement.
    \item $c_n = m$
    \item Every object in the chain pushes against the next object in the chain: $Z_{c_i}(p_{c_i} + \hat{u}_a) \cap Z_{c_{i+1}}(p_{c_{i+1}}) \neq \emptyset$ for $i = 1, \dotsb, n-1$.
\end{itemize}
The wall object $W$ cannot change position, so if $M$ contains $W$, then no objects move, resulting in $s' = s$. If not, then:

$$
p_i'= \begin{cases}
  p_i + \hat{u}_a & \textrm{if}\ i \in M \\
  p_i & \textrm{otherwise}
\end{cases}
$$
for every object $i$. $T$ copies the object cells $Y_i$ without modification.

In one variation, the agent object may be constrained by additional walls that all other objects can move through. These agent-specific walls can force the agent to use tools to manipulate objects out of reach.

\subsection{Classical Planner Implementations}
\label{appendix.classical_planners}

We used the following implementations of these classical planners:

\begin{itemize}

    \item \textbf{Fast Forward}:  We used the version included in the Fast Downward project at \url{https://www.fast-downward.org/}. 

    \item \textbf{Fast Downward}: We used the ``seq-sat-fd-autotune-1" version included in the Fast Downward project, and this version uses the context-enhanced additive heuristic \cite{helmert2008unifying} that succeeded the causal graph heuristic. This version outperforms ``seq-sat-fd-autotune-2" in \pw.
    
    \item \textbf{LAMA}: We used the ``seq-sat-lama-2011" version included in the Fast Downward project.
    
    \item \textbf{Best-First Width Search}: We used the version from \url{https://github.com/nirlipo/BFWS-public} and only report results for the ``DUAL-BFWS" mode, which outperforms ``BFWS-f5" and ``k-BFWS" in \pw.
    
    \item \textbf{Fast Downward Stone Soup}: We used the ``seq-sat-fdss-2018" version included in the Fast Downward project.
    
    \item \textbf{Saarplan}: We used the version from IPC 2018 at \url{https://ipc2018-classical.bitbucket.io/}.

\end{itemize}

\subsection{Recursive Graph Distance Pseudocode}
\label{appendix.rgd}

Algorithm~\ref{rgd} defines the RGD heuristic in the context of \pw. This implementation uses \textit{movement graphs}, which are equivalent to object position DTGs that omit parallel edges. Instead of labeling graph edges with operator preconditions, the implementation uses the \textproc{RelativePushingPositions} function to return all relative position offsets from which one object can push another in a given direction based on the objects' shapes. The movement graphs omit all positions of objects that would collide with walls, so the implementation does not include explicit collision checks.

In the reported experimental results, the RGD implementation includes the following optimizations, which are not shown in Algorithm~\ref{rgd}:

\begin{itemize}
    \item The heuristic ignores negative preconditions of operators; it does not check for whether movable obstacles would block a movement. Such a check could be exponentially expensive in the number of objects because the movement might be blocked by an obstacle that is blocked by another obstacle, etc., which is finally blocked by a wall. This choice does not affect the completeness of a search that uses this heuristic.
    \item Instead of allowing unbounded recursion depth in \textproc{PushingCost},
    which can be exponentially expensive in the number of objects, planning time is empirically reduced by incrementally increasing the maximum recursion depth until the heuristic cost is not infinite.
    \item The calculation of $d_\text{min}$ within the \textbf{for} loop on line 30 is cached for all pairs of objects in all positions.
    \item \textproc{ShortestPathLength} computes the length of the shortest directed path between two nodes in a graph. These path lengths are cached, and computing the shortest path length between one pair of nodes can reduce the computation in subsequent calls for different nodes. One possible implementation is to maintain a breadth-first expansion from each target node and to continue the expansion whenever a source node is not in the visited set whose shortest path lengths to the target are known.
    \item \textproc{RelativePushingPositions} is memoized to avoid repeated comparisons of the same object shapes.
\end{itemize}

\begin{algorithm*}
\caption{Recursive Graph Distance Heuristic}
\label{rgd}
\begin{algorithmic}[1]

\Function{RecursiveGraphDistanceCost}{$\text{state}, \text{goal}, \text{movement\_graphs}$}
    \State $c \gets 0$
    \ForAll {$(\text{object\_id}, p_\text{goal}) \in \text{goal}$}
        \State $c \gets c + \Call{CostToReachPosition}{
            \text{object\_id}, p_\text{goal}, \text{state}, \text{movement\_graphs}}$
        \If {$c = \infty$}
            \State \textbf{break}
        \EndIf
    \EndFor
    \State \textbf{return} c
\EndFunction

\item[]

\Function{CostToReachPosition}{$\text{object\_id}, p_\text{goal}, \text{state}, \text{ movement\_graphs}$}
    \State $p \gets \text{state}[\text{object\_id}].\text{position}$
    \If {$p = p_\text{goal}$}
        \State \textbf{return} $0$
    \EndIf
    \State $c_\text{min} \gets \infty$
    \State $\text{used\_object\_ids} \gets \{\text{object\_id}\}$
    \ForAll {$p_\text{next} \in \text{movement\_graphs}[\text{object\_id}].\text{direct\_successors}(p)$}
        \State $d \gets \Call{ShortestPathLength}{
            \text{movement\_graphs}[\text{object\_id}], 
            p_\text{next}, 
            p_\text{goal}}$ \Comment{Returns $\infty$ if no path exists.}
        \If {$d < c_\text{min}$} 
            \State $c_\text{min} \gets d + \Call{PushingCost}{
                \text{object\_id}, p_\text{next}, \text{used\_object\_ids}, \text{state}, \text{movement\_graphs}, c_\text{min} - d}$
        \EndIf
    \EndFor
    \State \textbf{return} $c_\text{min}$
\EndFunction

\item[]

\Function{PushingCost}{$\text{object\_id}, p_\text{next}, \text{used\_object\_ids}, \text{state}, \text{movement\_graphs}$, \text{cost\_upper\_bound}}
    \State $c_\text{min} \gets \text{cost\_upper\_bound}$
    \State $p \gets \text{state}[\text{object\_id}].\text{position}$
    \State $\hat{u}_a \gets p_\text{next} - p$

    \For {$\text{pusher\_object\_id} \gets 1 \textbf{ to } |\text{state}|$}
        \If {$\text{pusher\_object\_id} \in \text{used\_object\_ids}$}
            \State \textbf{continue}
        \EndIf

        \State $p^\text{pusher} \gets \text{state}[\text{pusher\_object\_id}].\text{position}$
        \State $\text{next\_used\_object\_ids} \gets \text{used\_object\_ids} \cup \{\text{pusher\_object\_id}\}$

        \LineComment{For all pusher positions that are adjacent to the pusher's current position, compute the minimum graph}
        \LineComment{distance from each adjacent position to the position where the pusher makes contact with the object\_id.}
        
        \ForAll {$p_\text{next}^\text{pusher} \in \text{movement\_graphs}[\text{pusher\_object\_id}].\text{direct\_successors}(p^\text{pusher})$}
            \State $d_\text{min} \gets \infty$

            \ForAll {$\Delta p \in \Call{RelativePushingPositions}{
                    \text{state}[\text{object\_id}].\text{shape}, \text{state}[\text{pusher\_object\_id}].\text{shape}, \hat{u}_a}$}
                \State $p_\text{start}^\text{pusher} \gets p + \Delta p$
                \State $p_\text{end}^\text{pusher} \gets p_\text{start}^\text{pusher} + \hat{u}_a$

                \If {$( p_\text{start}^\text{pusher}, p_\text{end}^\text{pusher} ) \notin \text{movement\_graphs}[\text{pusher\_object\_id}].\text{edges}$}
                    \State \textbf{continue}
                \EndIf

                \If {$p_\text{start}^\text{pusher} = p^\text{pusher} \textbf{ and }
                        p_\text{end}^\text{pusher} = p_\text{next}^\text{pusher}$}
                    \Comment{This is a simultaneous push, so there is no cost.}
                    \State $d_\text{min} \gets 0$
                    \State \textbf{break}
                \Else
                    \State $d \gets \Call{ShortestPathLength}{
                        \text{movement\_graphs}[\text{pusher\_object\_id}],
                        p_\text{next}^\text{pusher}, p_\text{start}^\text{pusher}
                    }$
                    \State $d \gets d + 1$ \Comment{Add the cost of moving the pusher from its start position to its end position.}
                \EndIf

                \State $d_\text{min} \gets \text{min}(d_\text{min}, d)$
            \EndFor
            
            \If {$\text{pusher\_id} = \text{AGENT\_OBJECT\_ID}$}
                \State $c_\text{min} \gets \text{min}(c_\text{min}, d_\text{min} + 1)$ \Comment{Moving the agent to its next position costs 1 action.}
            \ElsIf {$d_\text{min} < c_\text{min}$}
                \State \begin{varwidth}[t]{\linewidth}
                  $c_\text{min} \gets d_\text{min} + \Call{PushingCost}{$\par \hskip\algorithmicindent
                    $\text{pusher\_object\_id}, p_\text{next}^\text{pusher},
                    \text{next\_used\_object\_ids}, \text{state}, \text{movement\_graphs}, c_\text{min} - d_\text{min}}$
                    \end{varwidth}
            \EndIf
        \EndFor
    \EndFor
    \State \textbf{return} $c_\text{min}$
\EndFunction
\end{algorithmic}
\end{algorithm*}

\subsection{Deep RL Network Architecture and Hyperparameters}
\label{appendix.network_architecture}

We used the PPO and DQN implementations in DeepMind Acme \cite{hoffman2020acme}: \url{https://github.com/deepmind/acme}. For both PPO and DQN, we used a convolutional neural network (CNN) \cite{726791} as the vision processor, implemented in JAX (\url{https://github.com/google/jax}). The CNN consisted of 3 convolutional layers below 2 fully connected layers, and every layer used ReLU activation. The three convolutional layers used filter sizes of 3x3, 3x3, and 5x5, with stride lengths of 3, 1, and 1. The fully connected layers had sizes 256 and 128, respectively.

The PPO hyperparameters were: entropy cost 0.01, learning rate 0.0002, and number of epochs 2. The DQN hyperparameters were: learning rate 0.0001, epsilon 0.05, samples per insert 2, batch size 256, discount 1.0, and 1-step updates. These parameters were determined via hyperparameter sweeps.

We set the episode length to 100 steps, considering that Level 0 puzzles can be solved in fewer than 15 steps on average, Level 1 puzzles can be solved in fewer than 30 steps on average, and puzzles with multiple goals can provide positive rewards without solving the entire puzzle.

\subsection{PPO and DQN Results}
\label{appendix.ppo_dqn_train_test_results}

\begin{figure*}[htbp!]
\centering
\includegraphics[width=0.9\textwidth]{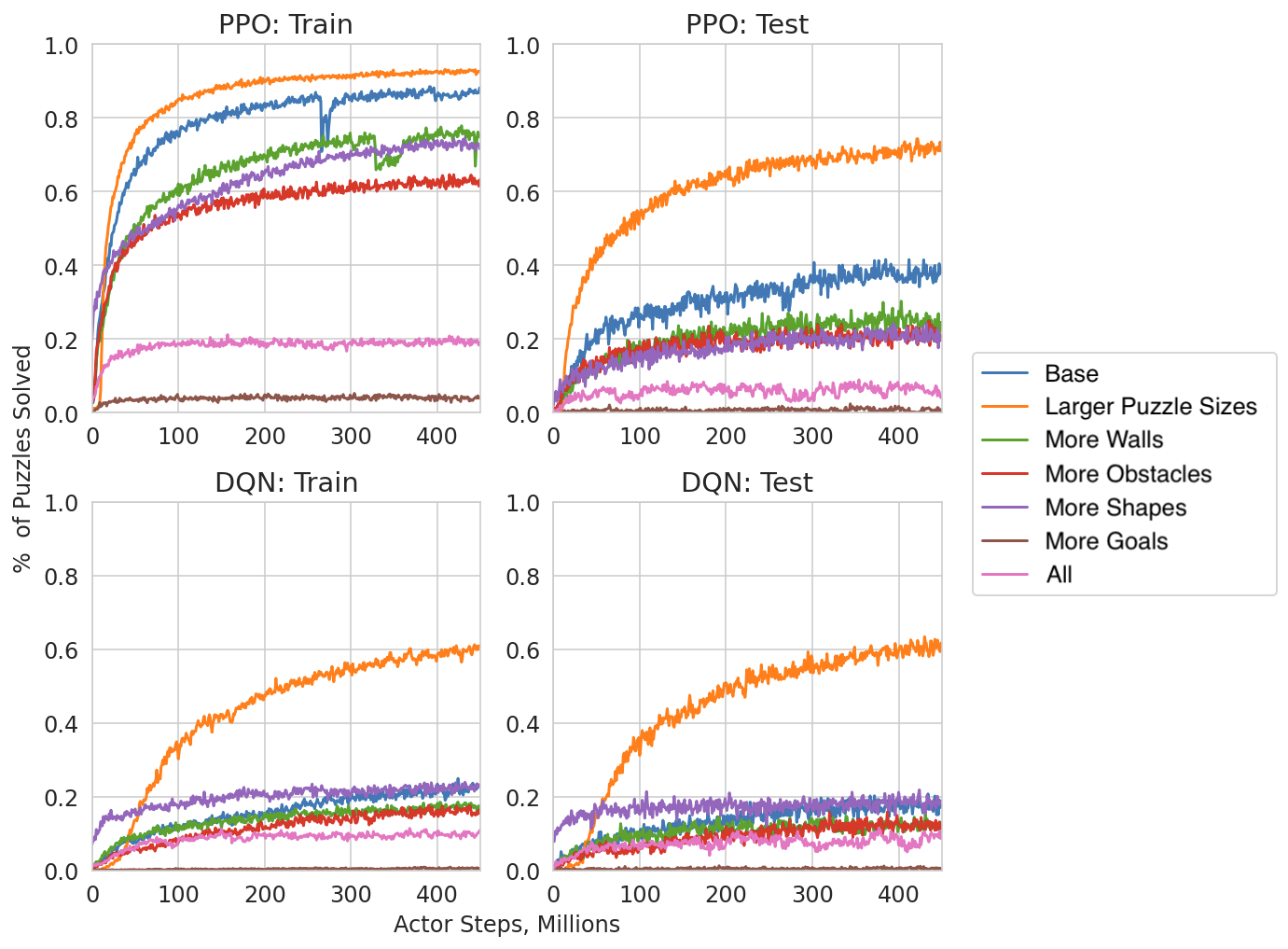}
\caption{PPO and DQN training and test curves on the Level 0 puzzle sets.}
\label{fig:pushworld_results_rl_combined}
\end{figure*}

Figure \ref{fig:pushworld_results_rl_combined} shows the performance of PPO and DQN on Level 0 puzzle sets.

\end{document}